\documentclass[11pt]{article}
\usepackage{acl}
\usepackage{times}
\usepackage{latexsym}
\usepackage[T1]{fontenc}
\usepackage[utf8]{inputenc}
\usepackage{microtype}
\usepackage{inconsolata}
\usepackage{graphicx}
\usepackage{amsmath}
\usepackage{booktabs}
\usepackage{multirow}
\usepackage{array}
\usepackage{subcaption}
\usepackage{amssymb}
\usepackage{pifont}
\usepackage{tabularx}
\usepackage{float}

\title{LinguIUTics at PsyDefDetect: Iterative Imbalance-Aware Fine-tuning
of Qwen3-8B for Psychological Defense Mechanism Classification%
\thanks{Accepted at PsyDefDetect, a shared task at
the 25th BioNLP Workshop (BioNLP 2026), co-located with
ACL 2026 in San Diego, CA, USA.}}
\author{
\textbf{Shefayat E Shams Adib}$^*$,
\textbf{Ahmed Alfey Sani}$^*$,
\textbf{Md Hasibur Rahman Alif}$^*$, \\
\textbf{Ajwad Abrar}$^*$ \\[6pt]
Department of Computer Science and Engineering, \\
Islamic University of Technology, Dhaka, Bangladesh \\
\texttt{\{shefayatadib, ahmedalfey, hasiburrahman21, ajwadabrar\}@iut-dhaka.edu} \\
\small $^*$All authors contributed equally to this work.
}

\begin{document}

\maketitle

\begin{abstract}
Detecting psychological defense mechanisms in conversational text 
remains a challenging clinical NLP problem. For the PsyDefDetect 
2026 shared task (9-class utterance classification evaluated via 
macro F1), our team LinguIUTics\footnote{Code and resources available at
\url{https://github.com/Shefwef/LingIUTics-PsyDefDetect-BIONLP26}} achieves a macro F1-score of 0.3917 on the official positive-class leaderboard, ranking 4th out of 21 registered teams and
improving over the Ministral-8B task baseline (31.48 macro F1) by +7.7 absolute points (+24.4\% relative). BERT-family encoders and zero-shot LLMs proved ineffective on rare classes due to severe class imbalance, leading 
us to QLoRA fine-tuning of Qwen3-8B. We leverage three key 
strategies: grouped stratified cross-validation (preventing 
leakage), minority-class round-robin lexical augmentation, and a 
post-processing pipeline with logit bias tuning and ensemble 
blending. Together, these components close much of the validation–leaderboard gap and substantially improve minority-class recall, driving the
critical ``Unclear'' class (Level 8) from near-zero performance
to F1\,=\,0.797.
\end{abstract}

\section{Introduction}

Automatic detection of psychological defense mechanisms (unconscious strategies to mitigate distress under the DMRS framework \citep{perry2004dmrs}) helps mental health platforms flag maladaptive coping and improves empathetic conversational agents \citep{liu-etal-2021-towards, na-etal-2025-survey}. The PsyDefDetect 2026 shared task \citep{na-etal-2026-psydefdetect, na-etal-2026-psydefconv} challenges participating systems to classify seeker utterances into nine DMRS levels. This poses a major obstacle, that is extreme class imbalance \citep{garcia2009imbalanced}, with the frequency gap between majority ("High-Adaptive", 51.9\%) to minority ("Unclear ",1.5\% ) classes at about 34.6 times respectively. Because evaluation uses macro-averaged F1, optimizing on accuracy leads to majority-class collapse and task failure.

To address this, we followed up with an iterative development process. Standard single-fold PEFT (parameter-efficient fine-tuning) on Qwen3-8B \citep{yang2025qwen3} suffered a massive generalization gap (0.34 validation vs. 0.24 leaderboard F1) due to limited low-rank capacity and majority-class overfitting. By systematically upgrading model capacity, loss functions, and inference, we established a robust pipeline. Our key contributions are:
\begin{enumerate}
    \item A leakage-safe cross-validation scheme at the level of groups, where synthetic augmentations are in a set with their source utterances. This leads to an order of magnitude smaller generalization gap between out-of-fold and leaderboard batches. \item An oversampling method that preserves the original psychological signal\citep{wei2019eda}. This is achieved by expanding specific minority classes by 3 times in a round-robin lexical mutation approach. \item A post-processing pipeline that combines OOF based logit bias tuning, that is guarded using v2 decoding, and multi-seed probability blending.
\end{enumerate}

\section{Task and Dataset}

The PsyDefDetect 2026 task \citep{na-etal-2026-psydefdetect} 
classifies seeker utterances into nine psychological defense 
levels, as defined by the DMRS framework \citep{perry2004dmrs}, and evaluated by macro averaged F1-score. The PsyDefConv dataset 
\citep{na-etal-2026-psydefconv} contains 2,336 utterances across 
200 dialogues from ESConv \citep{liu-etal-2021-towards}. For 
5-fold CV, we merged train and validation sets (1,864 training 
examples and 472 test). It exhibits very high class imbalance 
with a 34.6 times frequency gap (Table~\ref{tab:data}).

\begin{table}[ht]
  \centering
  \small
  \resizebox{\columnwidth}{!}{%
  \begin{tabular}{clrr}
  \toprule
  \textbf{L} & \textbf{Defense Mechanism} & \textbf{N} & \textbf{\%} \\
  \midrule
  0 & No Defense / Neutral       & 296 & 15.9 \\
  1 & Action Defenses            & 108 & 5.8  \\
  2 & Major Image-Distorting     & 61  & 3.3  \\
  3 & Disavowal                  & 99  & 5.3  \\
  4 & Minor Image-Distorting     & 84  & 4.5  \\
  5 & Neurotic                   & 48  & 2.6  \\
  6 & Obsessional                & 172 & 9.2  \\
  \textbf{7} & \textbf{High-Adaptive}   & \textbf{968} & \textbf{51.9} \\
  8 & Unclear / Needs More Info  & 28  & 1.5  \\
  \midrule
    & \textbf{Combined Train}    & \textbf{1,864} & \textbf{100.0} \\
  \bottomrule
  \end{tabular}%
  }
  \caption{PsyDefConv combined training class distribution (Train + Val splits). Level 7 vs.\ Level 8 ratio: 34.6$\times$.}
  \label{tab:data}
\end{table}

\section{Implementation Process}
\label{sec:journey}

Our development went through three iterative stages, each revealing a 
fundamental limitation that directly inspired the next architectural 
transition. Table~\ref{tab:consolidated_results} follows the complete 
leaderboard path from R0 (F1\,=\,0.240) to our final submission 
(F1\,=\,0.392). The detailed system run log is provided in 
Table~\ref{tab:runs}.

\begin{table}[ht]
  \centering
  \small
  \resizebox{\columnwidth}{!}{%
  \begin{tabular}{lcc}
  \toprule
  \textbf{System / Variant} & \textbf{OOF F1} & \textbf{LB F1} \\
  \midrule
  R0: MentalBERT            & N/A & 0.240 \\
  R1: MentalBERT+RoBERTa ensemble       & N/A & 0.240 \\
  R2: MentalRoBERTa Focal + EMD   & 0.314 & N/A \\
  R3: DeBERTa-v3-base 5-fold   & 0.307 & 0.236 \\
  R4: RoBERTa-base               & N/A & 0.269 \\
  R5: Qwen3-8B 1-fold $r$=64 baseline   & 0.345 & 0.249 \\
  R6: Qwen3-8B 5-fold $r$=128 + weighted CE       & 0.361 & 0.329 \\
  R7: Qwen3-8B v2 & 0.372 & 0.355 \\
  R8: Qwen3-8B v2 microplus       & 0.372 & 0.354 \\
  R9: Qwen3-8B seed A only + v2 decode             & 0.431 & N/A \\
  \textbf{R10: Qwen3-8B old + seed A blend + v2 decode } & \textbf{0.437} & \textbf{0.392} \\
  \bottomrule
  \end{tabular}%
  }
  \caption{Complete system run log. LB = CodaBench leaderboard.}
  \label{tab:runs}
\end{table}

\begin{figure*}[t]
  \centering
  \includegraphics[width=\textwidth]{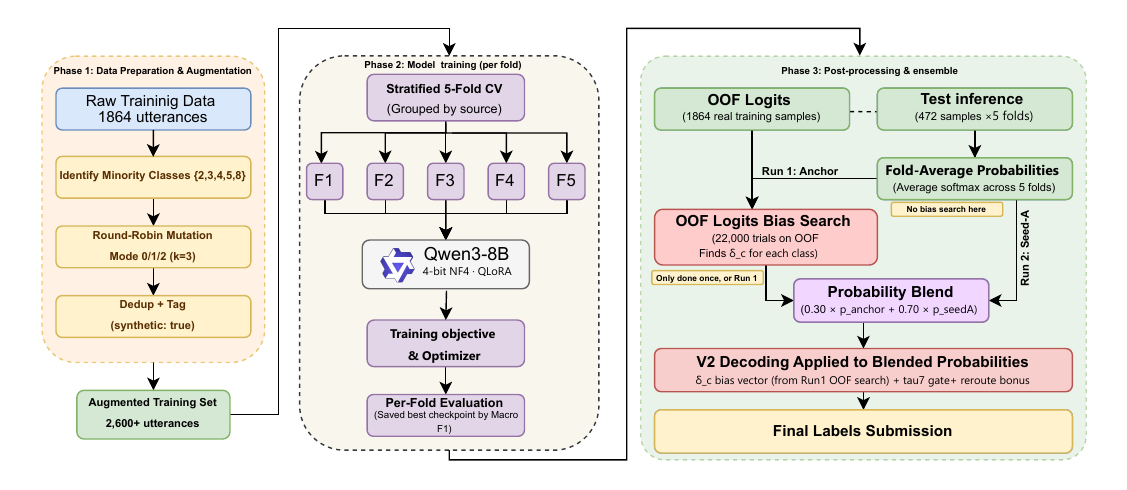}
\caption{%
  Full system pipeline.
  \textbf{Phase~1}: raw training data is preprocessed and
  minority classes (Levels 2, 3, 4, 5, 8) are oversampled via
  round-robin lexical mutation, creating an expanded training set
  of 2,600+ utterances.
  \textbf{Phase~2}: two independent grouped stratified 5-fold
  QLoRA fine-tuning runs of Qwen3-8B the \textit{Anchor}
  (seed\,=\,42) and \textit{Seed-A} (seed\,=\,20260407) sharing identical architecture and hyperparameters, with
  dialogues grouped across folds to prevent data leakage.
  \textbf{Phase~3}: a class-specific logit bias vector
  ($\delta_c$) is optimised on Anchor OOF predictions and locked
  for reuse; test probabilities from both runs are blended before
  a final guarded decode step that maximises minority-class recall
  without sacrificing majority-class precision.
}
  \label{fig:pipeline}
\end{figure*}
\subsection{Stage 1: BERT-Family Encoder Baselines}
\label{sec:stage1}

We evaluated MentalBERT, MentalRoBERTa, 
DeBERTa-v3-base, and RoBERTa-base 
\citep{devlin2019bert, ji-etal-2022-mentalbert, he2021deberta, liu2019roberta} 
for multiple context windows, loss functions (cross-entropy, Focal 
\citep{lin2017focal}, EMD), and ensemble strategies (full scores in 
Appendix~\ref{app:allmodels}, Table~\ref{tab:consolidated_results}). 
Best validation macro F1 was 0.314 (MentalRoBERTa + Focal/EMD 
+ Hungarian remapping), with leaderboard peak at 0.240. Importantly, 
F1 for Classes~3, 5 and~8 was still zero across \emph{all} variants, 
establishing an \emph{encoder capacity ceiling} with a 51.8\% 
majority-class prior using $n\!\leq\!50$ minority examples. From the general formula:
{
 \setlength{\abovedisplayskip}{4pt}
 \setlength{\belowdisplayskip}{4pt}
\begin{equation}
  \mathcal{L}_{\text{Focal}} = -(1-p_t)^{\gamma} \log p_t
  \label{eq:focal}
\end{equation}
}%
where $p_t$ is the probability of predicting the correct class, 
logarithmically compensated by strength constant $\gamma$.

\subsection{Stage 2: Zero-Shot Evaluation}
\label{sec:stage2}
Qwen3-8B, Llama~3.1-8B, and Ministral-8B evaluated zero-shot with 
explicit DMRS label definitions produced 8--16\% macro F1 (Table~\ref{tab:consolidated_results}), confirming that task knowledge cannot be prompt-engineered.

\subsection{Stage 3: Diagnostic LLM Fine-Tuning}
\label{sec:stage3}
Ministral-8B fine-tuned with 4-bit NF4 quantization achieved 64.71\% 
accuracy but only 14.74 macro F1 (Table~\ref{tab:consolidated_results}). 
This illustrates that standard cross-entropy collapses to majority-class 
prediction under severe imbalance. Furthermore, the accuracy is an 
actively misleading metric in this setting.
\subsection{Stage 4: Final System (Qwen3-8B LoRA Pipeline)}
\label{sec:stage4}

As such, with the three lessons above guiding us, our final pipeline 
consists of five components: model architecture, imbalance-aware 
training objective, data augmentation, leakage-safe cross-validation and a post-processing ensemble, each targeting a specific failure mode identified in the earlier stages.

\subsubsection{Model Architecture}

We fine-tune Qwen3-8B via QLoRA \citep{dettmers2023qlora} with 4-bit 
NF4 quantization, reducing peak GPU memory from ${\sim}$32\,GB to 
${\sim}$8\,GB on a single NVIDIA RTX 3090 Ti. LoRA adapters 
\citep{hu2022lora} target all attention and feed-forward layers 
(\texttt{q}, \texttt{k}, \texttt{v}, \texttt{o}, \texttt{gate}, 
\texttt{up}, \texttt{down}) and the \texttt{score} head 
($r\!=\!128$, $\alpha\!=\!256$, dropout\,=\,0.1), yielding 
${\approx}$31M trainable parameters (0.4\% of the 8B base). 
Increasing rank from $r\!=\!64$ to $r\!=\!128$ delivered a 
$+$24.9\% fold-level F1 gain, critical for separating 
psychologically similar classes (e.g., Level~4 vs.\ Level~5).
Full training hyperparameters for this configuration are summarised in Table~\ref{tab:hyper} in Appendix~\ref{app:hyper}.

\subsubsection{Input Representation}
Each prompt has three parts: (1)~the DMRS Label Guide with 
9-class clinical schema; (2)~Conversational Context of the last 30 dialogue turns prefixed by \texttt{SEEKER:}/\texttt{HELPER:} tags and (3)~an Output Instruction directing the model to emit a single integer ($0$--$8$). Inputs are tokenised to 1,024 tokens with dynamic padding to multiples of 8, covering $>$95\% of samples without truncation.

\subsubsection{Imbalance-Aware Training Objective}
This dataset is, to a large extent, imbalanced towards Level~7. To 
mitigate the problem of majority-class collapse, we apply two 
stabilization techniques.

\paragraph{Inverse-Square-Root Class Weighting.}
Per-class weights
{\setlength{\abovedisplayskip}{4pt}
 \setlength{\belowdisplayskip}{4pt}
\begin{equation}
\label{eq:weights}
w_c = \tfrac{1/\sqrt{n_c}}{\sum_{i=0}^{8}1/\sqrt{n_i}}
\end{equation}
}boost the most under-represented classes (e.g., $w_8\!=\!1.67$, 
$w_5\!=\!1.29$) while dampening the majority class ($w_7\!=\!0.28$), 
instead of inverse-frequency weighting that can lead to gradient 
instabilities.

\paragraph{Label Smoothing \& Optimization Schedule.}
By using label smoothing \citep{szegedy2016rethinking} 
($\varepsilon\!=\!0.05$), we avoid early logit saturation for 
Level~7. This prevents gradients from being dominated by the majority
class and allows minority-class signals to contribute more effectively
during back-propagation. We employ AdamW with a peak learning rate of $1.2\!\times\!10^{-4}$, cosine 
annealing (8\% warmup), 10 epochs per fold, batch size 16 
($2\!\times\!8$ accumulation), and gradient clipping at 0.3 
(see Appendix~\ref{app:hyper}).

\subsubsection{Data Augmentation}
For the rarest classes (Levels 2, 3, 4, 5, and 8), there are between 
28 and 84 training examples. This is not enough for an 8B capacity 
model to learn sufficiently reliable decision boundaries. In order to 
fix this, we perform \textbf{round-robin lexical mutation} to generate 
$k=3$ synthetic variants per source utterance in these classes, cycling 
through three surface-level rewriting modes:

\begin{itemize}
    \setlength{\itemsep}{2pt}
    \setlength{\parskip}{0pt}
    \item \textbf{Mode 0:} Contraction replacement 
    (e.g., \textit{I am} $\to$ \textit{I'm}) plus a hedging prefix 
    (\textit{Honestly, \ldots}).

    \item \textbf{Mode 1:} Vocabulary style-shift 
    (e.g., \textit{maybe} $\to$ \textit{perhaps}) plus a trailing 
    filler (\textit{\ldots\ I guess.}).

    \item \textbf{Mode 2:} Hesitation markers 
    (e.g., periods $\to$ ellipses; \texttt{?} $\to$ \texttt{??}).
\end{itemize}
Mutations target only the seeker utterance to preserve the 
psychological signal; after deduplication, minority class 
counts increased from 28--84 to 65--252 examples 
(see Appendix~\ref{app:aug}). This targeted minority-class oversampling strategy is consistent with prior findings that augmenting only underrepresented classes yields more effective and stable performance improvements than augmenting all classes equally \cite{sani2026addressing}.

\subsubsection{Grouped Five-Fold Cross-Validation}

Random splitting risks leakage across the 200 source dialogues, 
making dialogue-level isolation essential. We therefore apply 
grouped stratified cross-validation (grouped CV, implemented as \texttt{StratifiedGroupKFold} with $k=5$) using \texttt{dialogue\_id} as the grouping key:
\begin{itemize}
    \item \textbf{Zero Leakage Guarantee} (0 leaked dialogues confirmed): 
    All utterances and their synthetic variants are kept entirely within 
    one fold.
    \item \textbf{Reliable Validation Signal} (OOF leaderboard gap 
    reduced from 9.6 to 1.7-4.5 points): Strong rank-correlation 
    between OOF and leaderboard gains enables safe threshold tuning.
    \item \textbf{Ensemble Foundation} (5 checkpoints per seed): Five-fold 
    training yields pure OOF predictions for post-processing calibration 
    and reduces inference variance.
\end{itemize}
Reliable validation behaviour across folds is shown in the per-fold OOF metrics in Table~\ref{tab:folds}.

\begin{table}[H]
  \centering
  \small
  \begin{tabular}{lcccc}
  \toprule
  \textbf{Fold} & \textbf{Acc.} & \textbf{Mac.\ F1} & \textbf{Mac.\ P} & \textbf{Mac.\ R} \\
  \midrule
  1             & 0.6193 & 0.3804 & 0.3836 & 0.3925 \\
  2             & 0.6247 & 0.3701 & 0.3902 & 0.3617 \\
  3             & 0.6408 & 0.3899 & 0.4276 & 0.4000 \\
  4             & 0.6300 & 0.3553 & 0.3753 & 0.3514 \\
  5             & 0.5968 & 0.3326 & 0.3396 & 0.3298 \\
  \midrule
  \textbf{OOF} & \textbf{0.6223} & \textbf{0.3716} & \textbf{0.3817} & \textbf{0.3675} \\
  \bottomrule
  \end{tabular}
  \caption{Per-fold CV results (grouped-clean augmented run, before seed blending).}
  \label{tab:folds}
\end{table}

\subsubsection{Post-Processing and Ensemble Strategy}
Despite class-weighted training, raw probabilities remain heavily 
biased towards the majority class (Level~7). To rectify this and 
recover rare classes without compromising precision, we implement 
a three-stage post-training pipeline (\texttt{v2decode}).

\paragraph{Stage A: OOF Bias Optimization.}
Using logit adjustment for long-tail learning 
\citep{menon2021longtail}, we search for class-specific probability 
offsets ($\delta_c$) that maximize the OOF macro F1 score:
{
\setlength{\abovedisplayskip}{4pt}
\setlength{\belowdisplayskip}{4pt}
\begin{equation}
    \hat{y} = \arg\max_c \bigl[\log p_c + \delta_c\bigr]
\end{equation}
}
We evaluate approximately 22{,}000 randomly sampled bias vectors on
OOF predictions to identify a configuration that balances majority
precision with minority recall. The best locked vector applies a
negative penalty to Level~7 ($\delta_7 < 0$) and substantial positive bonuses to minority classes like Level~8 ($\delta_8 > 0$).

\paragraph{Stage B: Multi-Seed Blending.}
We run a second identical 5-fold training pipeline, denoted \textit{Seed-A}, using a different random seed (\texttt{seed = 20260407}) and the same architecture and hyperparameters. We combine the test-set probabilities of 
the original Anchor and Seed-A using a 30/70 weighted average:
{
\setlength{\abovedisplayskip}{4pt}
\setlength{\belowdisplayskip}{4pt}
\begin{equation}
    p_{\text{blend}} = 0.30 \cdot p_{\text{anchor}} + 
    0.70 \cdot p_{\text{seedA}}
\end{equation}
}
The ratio was tuned using real-only OOF F1, combining the 
Anchor's high precision with Seed-A's strong minority recall.

\paragraph{Stage C: The $\tau_7$-Gate Decoding.}
A confidence safeguard is applied from the locked $\delta_c$ bias 
vector onto $p_{\text{blend}}$ to prevent precision collapse:
\begin{itemize}
    \item \textbf{$\tau_7$-Protection Gate:} The prediction is 
    locked to Level~7 and $\delta_c$ offsets are not applied if 
    $p_{\text{blend},7} \ge 0.69$.
    \item \textbf{Minority Rerouting:} If $p_{\text{blend},7} 
    < 0.69$, $\delta_c$ offsets are applied, rerouting ambiguous 
    samples into the highest-probability minority class.
\end{itemize}
So, minority labels are aggressively recovered when the model is uncertain. This increases the minority recall without affecting the precision.
 
\section{Experiments and Results}
\label{sec:results}

\subsection{Cross-Paradigm Comparison}

BERT-family encoders struggled to break above 0.314 macro F1 due to 
capacity limits, across three paradigms 
(Table~\ref{tab:consolidated_results} in Appendix~\ref{app:allmodels}). 
Both zero-shot LLMs and standard LLM fine-tuning collapsed to 
majority-class predictions (near 15\% F1). In contrast, our 
imbalance-aware Qwen3-8B pipeline resolved these issues, reaching 
39.17\% macro F1.

\subsection{Comparison with SOTA Baselines}

Table~\ref{tab:leaderboard} compares the results of our systems 
against the task baselines 
\citep{na-etal-2026-psydefdetect, na-etal-2026-psydefconv}. 
Our final pipeline surpassed the stated SOTA, Ministral-8B fine-tuned baseline (31.48 macro F1) +7.7 absolute points, corresponding to a +24.4\% relative improvement in macro F1.

\begin{table}[ht]
\centering
\resizebox{\columnwidth}{!}{%
\begin{tabular}{lcc}
\toprule
\textbf{System} & \textbf{Acc. (\%)} & \textbf{Macro F1 (\%)} \\
\midrule
GPT-5 zero-shot (task paper)             & 52.75 & 19.53 \\
Gemini 2.5 Pro zero-shot                 & 56.36 & 25.99 \\
DeepSeek-V3.2 zero-shot (CoT)            & 55.72 & 26.17 \\
Llama 3.1-8B fine-tuned                  & 62.92 & 30.51 \\
InternLM3-8B fine-tuned                  & 63.98 & 30.53 \\
Ministral-8B fine-tuned (SOTA)           & 64.83 & 31.48 \\
\midrule
DeBERTa-v3-base (5-fold)                 & 59.11 & 23.58 \\
RoBERTa-base                             & 51.27 & 26.97 \\
Qwen3-8B LoRA (Baseline Finetuned)                 & 54.45 & 24.91 \\
Qwen3-8B LoRA (Grouped CV + Bias Tuning) & 58.43 & 35.48 \\
\textbf{Qwen3-8B LoRA (SeedA Ensemble + v2decode)} 
                                & \textbf{64.19} & \textbf{39.17}\\
                                
\bottomrule
\end{tabular}%
}
\caption{Comparison with task paper baselines.}
\label{tab:leaderboard}
\end{table}

\subsection{Ablation Study}

Table~\ref{tab:ablation} analyses each 
component's contribution. Increasing LoRA rank to $r\!=\!128$ produced the \linebreak 
\begin{table}[ht]
  \centering
  \small
  \begin{tabular}{lc}
  \toprule
  \textbf{System Configuration} & \textbf{Macro F1} \\
  \midrule
  R0: 1-fold, $r$=64, no weighting   & 0.249 \\
  + 5-fold CV, $r$=128               & 0.284$^{\dagger}$ \\
  + Weighted CE + label smoothing    & 0.329$^{\dagger}$ \\
  + Grouped-clean 5-fold             & 0.355 \\
  + Data augmentation (RR-k3)        & 0.355 \\
  + Seed-A blend (30/70) + v2 decode  & \textbf{0.392} \\
  \bottomrule
  \end{tabular}
  \caption{Ablation: each component's contribution. $^{\dagger}$Metrics for these rows are single-fold estimates from the 5-fold setup, included as indicative rather than full OOF results.}
  \label{tab:ablation}
\end{table}
\noindent
highest boost ($+$24.9\% fold-level F1), supporting that model capacity was indeed the primary bottleneck. Grouped CV, data augmentation, and 
post-processing decode rules contributed incrementally, securing the 
final $+$3.69 F1 points.

\subsection{Per-Class Analysis}

Per-class performance is shown in Figure~\ref{fig:cm_stacked} 
(Appendix~\ref{app:cm}) and Table~\ref{tab:perlabel_copy}. Level~8 (``Unclear'') saw the most 
improvement, climbing from near-zero recall to 0.797 F1 via 
augmentation and bias tuning, with Levels~2 and~3 also gaining 
substantially. On the other hand, Levels~4 and~5 continue to be 
difficult (0.25--0.27 F1) due to high linguistic overlap with the 
majority class. Importantly, optimising for minority classes did not 
compromise the majority class (Level~7), which still resulted in a 
solid F1 of 0.709.
 
\begin{table}[ht]
\centering
\small
\begin{tabular}{llccc}
\toprule
\textbf{L} & \textbf{Mechanism} & \textbf{P} & \textbf{R} & \textbf{F1} \\
\midrule
0 & Neutral        & 0.747 & 0.858 & 0.799 \\
1 & Action         & 0.242 & 0.398 & 0.301 \\
2 & Major Img-D    & 0.480 & 0.451 & 0.465 \\
3 & Disavowal      & 0.401 & 0.402 & 0.401 \\
4 & Minor Img-D    & 0.317 & 0.211 & 0.254 \\
5 & Neurotic       & 0.398 & 0.214 & 0.278 \\
6 & Obsessional    & 0.203 & 0.267 & 0.231 \\
7 & High-Adaptive  & 0.693 & 0.726 & 0.709 \\
8 & Unclear        & 0.797 & 0.797 & 0.797 \\
\midrule
\textbf{Macro} & & \textbf{0.431} & \textbf{0.436} & \textbf{0.426} \\
\bottomrule
\end{tabular}
\caption{Per-label OOF metrics, final blended system with v2 decode.
Level~8 improved from $\approx$0 to 0.797 via augmentation and bias
tuning.}
\label{tab:perlabel_copy}
\end{table}
 
\section{Conclusion}
\label{sec:conclusion}
We showed that the data-centric imbalance mitigation methods (grouped CV, weighted loss, round-robin lexical augmentation, and dynamic OOF bias tuning with ensembling) that we used were much more important than raw model capacity for psychological defense classification. We achieved a macro F1-score of 0.3917 on the official positive-class leaderboard, ranking 4th out of 21 registered teams. This corresponds to a +7.7 macro F1-score improvement (+24.4\% relative) over the Ministral-8B fine-tuned baseline. In future, we plan to add more effective paraphrase-based data augmentation, use losses better suited to imbalanced classes, and evaluate on more datasets.

\section*{Limitations}

These decode rules and OOF bias vectors are calibrated specifically to this dataset and so requires recomputation for unseen domains. The grouped CV protocol keeps mutant variants within the same source group to reduce leakage, but the risk of leakage cannot be fully eliminated. Lastly, we were limited to PEFT on models
with 8B parameters or less and by hardware constraints (24\,GB VRAM).

\section*{Acknowledgments}
 
We thank the PsyDefDetect 2026 shared task organizers 
\citep{na-etal-2026-psydefdetect} for providing the PsyDefConv dataset 
\citep{na-etal-2026-psydefconv} and evaluation infrastructure through 
CodaBench.

\bibliography{custom}
\newpage
\appendix

\section{Full DMRS Label Definitions}
\label{app:dmrs}
Table~\ref{tab:dmrs} provides the complete clinical descriptions for all nine psychological defense levels used in our classification prompt.

\begin{table*}[t]
  \centering
  \small
  \begin{tabularx}{\textwidth}{p{0.6cm}p{2.5cm}X}
  \toprule
  \textbf{L} & \textbf{Name} & \textbf{Clinical Description} \\
  \midrule
  0 & No Defense & Phatic or factual exchange with no active defense mechanism \\
  1 & Action & Emotional discharge through behavior (passive aggression, complaining, impulsive action) \\
  2 & Major Img-D & Extreme cognitive distortion: projection, splitting, all-or-nothing thinking \\
  3 & Disavowal & Avoiding unpleasant reality: denial, rationalization, minimization \\
  4 & Minor Img-D & Subtler distortion: devaluation, idealization, omnipotence \\
  5 & Neurotic & Unconscious anxiety management: repression, displacement, reaction formation \\
  6 & Obsessional & Over-intellectualization, isolation of affect, undoing \\
  7 & High-Adaptive & Mature coping: humor, altruism, insight, self-assertion, sublimation \\
  8 & Unclear & Insufficient context for reliable DMRS classification \\
  \bottomrule
  \end{tabularx}
  \caption{DMRS label definitions used in the classification prompt.}
  \label{tab:dmrs}
\end{table*}

\section{Full Hyperparameter Table}
\label{app:hyper}
\begin{table}[H]
  \centering
  \small
  \begin{tabular}{ll}
  \toprule
  \textbf{Hyperparameter} & \textbf{Value} \\
  \midrule
  Base model              & Qwen/Qwen3-8B \\
  Quantization            & 4-bit NF4 + double quant \\
  LoRA rank / alpha       & 128 / 256 \\
  LoRA dropout            & 0.1 \\
  LoRA target modules     & q/k/v/o/gate/up/down/score \\
  Trainable parameters    & $\approx$31M (0.4\%) \\
  Max sequence length     & 1024 \\
  Optimizer               & AdamW \\
  Learning rate           & $1.2 \times 10^{-4}$ \\
  Weight decay            & 0.01 \\
  LR scheduler            & Cosine annealing \\
  Warmup ratio            & 8\% \\
  Per-device batch size   & 2 \\
  Gradient accumulation   & 8 (eff.\ batch = 16) \\
  Gradient clip norm      & 0.3 \\
  Epochs per fold         & 10 \\
  Label smoothing $\varepsilon$ & 0.05 \\
  Class weight formula    & Inverse-sqrt (Eq.~\ref{eq:weights}) \\
  Hardware                & NVIDIA RTX 3090 Ti 24 GB \\
  Mixed precision         & bf16 \\
  \bottomrule
  \end{tabular}
  \caption{Complete training hyperparameters for the final Qwen3-8B LoRA system.}
  \label{tab:hyper}
\end{table}

\section{Data Augmentation Examples}
\label{app:aug}

Below are three mutation modes applied to a single Level-3 (Disavowal) sample. All mutations target only the seeker utterance; the supporting dialogue context is unchanged.

\begin{table*}[t]
\vspace{0.4cm}
\centering
\small
\begin{tabularx}{\textwidth}{>{\bfseries}p{2.2cm} X}
\toprule
\textbf{Mode} & \textbf{Utterance Text} \\
\midrule
Original \newline (Level 3) & \textit{``It is not really that bad honestly, I have been through worse situations before.''} \\
\addlinespace
Mode 0 \newline (Contractions \newline + Hedging) & \textit{``Honestly, it's not really that bad, I've been through worse situations before.''} \\
\addlinespace
Mode 1 \newline (Style Shift \newline + Filler) & \textit{``It is not quite that bad, I have been through worse situations before I guess.''} \\
\addlinespace
Mode 2 \newline (Hesitation) & \textit{``It is not really that bad honestly... I have been through worse situations before...''} \\
\bottomrule
\end{tabularx}
\caption{Example of round-robin lexical mutations for a Disavowal (Level-3) seeker utterance. The core signal (minimization (``not that bad'') and historical comparison (``been through worse'')) is preserved across all mutations, ensuring label validity.}
\label{tab:aug_examples}
\end{table*}

\section{Confusion Matrices}
\label{app:cm}

\begin{figure*}[t]
  \centering
  \begin{subfigure}{0.50\textwidth}
    \centering
    \includegraphics[width=\linewidth]{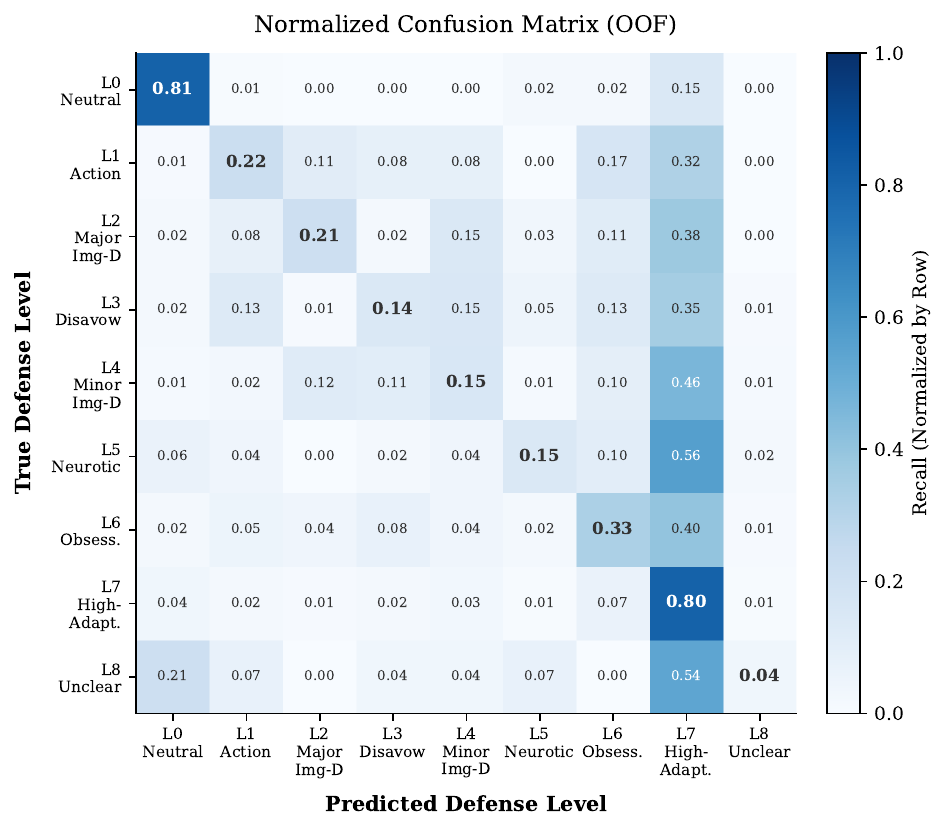}
    \vspace{-0.2cm}
    \caption{Before post processing}
    \label{fig:cm_before}
  \end{subfigure}\hfill
  \begin{subfigure}{0.50\textwidth}
    \centering
    \includegraphics[width=\linewidth]{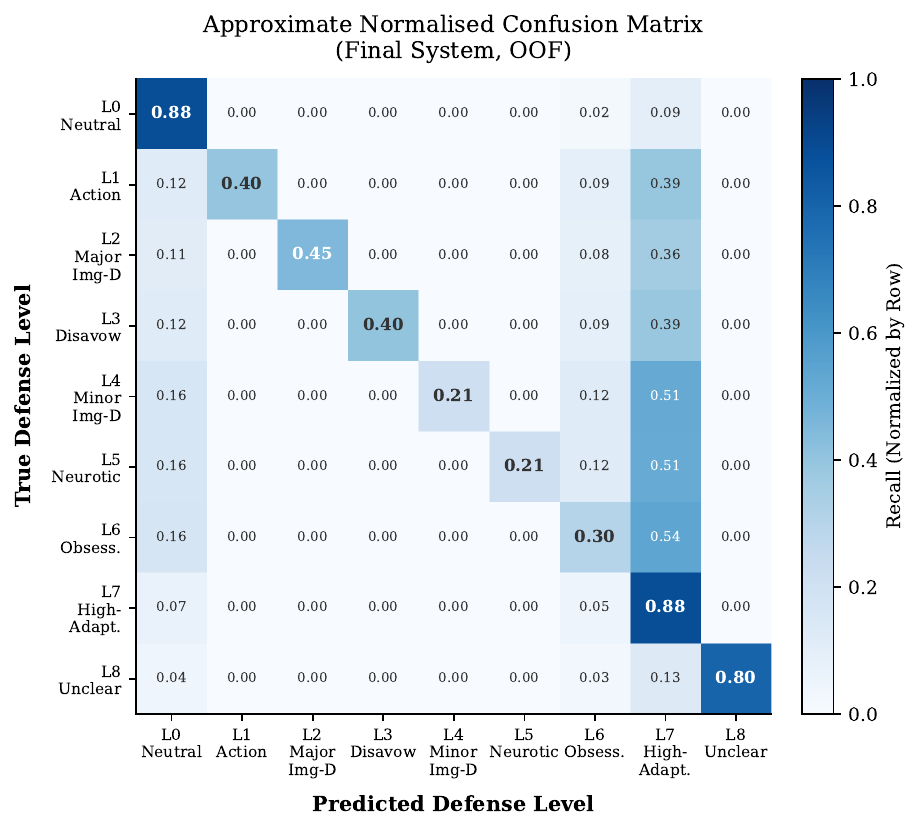}
    \vspace{-0.2cm}
    \caption{After post processing}
    \label{fig:cm_after}
  \end{subfigure}
  \caption{Approximate row-normalised confusion matrices (OOF). 
  Post-processing successfully shifts residual majority-class prediction bias away from L7, noticeably improving minority recall along the diagonal.}
  \label{fig:cm_stacked}
\end{figure*}

Figure~\ref{fig:cm_stacked} illustrates error distributions across minority and majority classes, highlighting grouping and filtering improvements.

\section{Comprehensive Model Comparison}
\label{app:allmodels}

Table~\ref{tab:consolidated_results} reports all systems evaluated during our development, organised by model family and experimental stage.
Empty cells indicate the model was not evaluated under that paradigm. Rows marked $\dagger$ are external baselines from the task paper \citep{na-etal-2026-psydefconv}; all others are our own internal tuning experiments.

\begin{table*}[t]
\vspace{0.4cm}
\centering
\resizebox{\textwidth}{!}{%
\begin{tabular}{l | cccc | cccc | cccc}
\toprule
\multirow{2}{*}{\textbf{Model}}
  & \multicolumn{4}{c|}{\textbf{BERT-family (Supervised)}}
  & \multicolumn{4}{c|}{\textbf{Zero-Shot LLM Inference}}
  & \multicolumn{4}{c}{\textbf{LLM Fine-Tuning (LoRA)}} \\
\cmidrule(lr){2-5} \cmidrule(lr){6-9} \cmidrule(lr){10-13}
  & Acc & macro F1 & Input / Loss & Ep
  & Acc & Macro Prec & Macro Recall & macro F1\,(1--8)
  & Acc & Precision & Recall & F1 \\
\midrule

MentalBERT-base
  & n/a & 0.240\,$\checkmark$ & Flat SEP $k$=5, Wtd CE, LR 2e-5 & 5
  & & & & & & & & \\[2pt]

MentalRoBERTa-base
  & n/a & 0.2200 & Flat SEP $k$=10, Wtd CE, LR 1e-5 & 10
  & & & & & & & & \\[2pt]

MentalBERT+RoBERTa (ens.)
  & n/a & 0.240\,$\checkmark$ & Flat SEP $k$=10, Wtd CE, LR 2e/1e-5 & 5/8
  & & & & & & & & \\[2pt]

DeBERTa-v3-base (5-fold)
  & 0.591 & 0.2358 & {[CTX]/[TGT]}, Wtd CE, LR 1.5e-5 & 8
  & & & & & & & & \\[2pt]

RoBERTa-base
  & 0.513 & 0.2697 & {[CTX]/[TGT]}, Wtd CE, LR 2e-5 & 6
  & & & & & & & & \\[2pt]

RoBERTa-base (len=320)
  & 0.479 & 0.2763 & {[CTX]/[TGT]}, Wtd CE, LR 1.5e-5 & 6
  & & & & & & & & \\[2pt]

RoBERTa-base + OS (unif.)
  & 0.618 & 0.2893 & {[CTX]/[TGT]}, Std CE + OS(120), LR 2e-5 & 6
  & & & & & & & & \\[2pt]

RoBERTa-base + OS (tgt.)
  & 0.614 & 0.2430 & {[CTX]/[TGT]}, Std CE + tgt.\ OS, LR 2e-5 & 8
  & & & & & & & & \\[4pt]

\midrule

Qwen3-8B
  & & & & 
  & 0.3422 & 0.1773 & 0.1670 & 0.1536
  & & & & \\[2pt]

Llama 3.1-8B
  & & & &
  & 0.3036 & 0.1958 & 0.1999 & 0.1584
  & & & & \\[2pt]

Ministral-8B
  & & & &
  & 0.2406 & 0.2517 & 0.1481 & 0.0841
  & & & & \\[4pt]

\midrule

Mistral-7B-v0.3 (Kaggle)
  & & & &
  & & & &
  & 0.1444 & n/a & n/a & 0.1023 (0.2054$^\dagger$) \\[2pt]

Ministral-8B (Local)
  & & & &
  & & & &
  & 0.6471 & n/a & n/a & 0.1474 \\[2pt]

\textbf{Qwen3-8B}\,$\bigstar$
  & & & &
  & & & &
  & \textbf{0.6419} & \textbf{0.4003} & \textbf{0.3958} & \textbf{0.3917}\,$\bigstar$ \\

\bottomrule
\end{tabular}%
}
\caption{%
  Consolidated results across all model families.
  Empty cells indicate the model was not evaluated under that paradigm.
  $\checkmark$ denotes selected ensemble members.
  $\bigstar$ denotes the best result.
  Rows marked $\dagger$ are external baselines from the task paper 
  \citep{na-etal-2026-psydefdetect}; all others are our own internal 
  tuning experiments.
}
\label{tab:consolidated_results}
\end{table*}

\end{document}